# Using the Tsetlin Machine to Learn Human-Interpretable Rules for High-Accuracy Text Categorization with Medical Applications

Geir Thore Berge[1,2,3], Ole-Christoffer Granmo[1], Tor Oddbjørn Tveit[1,3,4], Morten Goodwin[1*], Lei Jiao[1*], and Bernt Viggo Matheussen[1*]

[1] Centre for Artificial Intelligence Research, University of Agder, Grimstad, Norway
[2] Department of Information Systems, University of Agder, Kristiansand, Norway
[3] Department of Technology and eHealth, Sørlandet Hospital Trust, Kristiansand, Norway
[4] Department of Anesthesia and Intensive Care, Sørlandet Hospital Trust, Kristiansand, Norway
geir.thore.berge@sshf.no

## Abstract

Medical applications challenge today's text categorization techniques by demanding both high accuracy and ease-of-interpretation. Although deep learning has provided a leap ahead in accuracy, this leap comes at the sacrifice of interpretability. To address this accuracy-interpretability challenge, we here introduce, for the first time, a text categorization approach that leverages the recently introduced Tsetlin Machine. In all brevity, we represent the terms of a text as propositional variables. From these, we capture categories using simple propositional formulae, such as: **if** "rash" **and** "reaction" **and** "penicillin" **then** Allergy. The Tsetlin Machine learns these formulae from a labelled text, utilizing conjunctive clauses to represent the particular facets of each category. Indeed, even the absence of terms (negated features) can be used for categorization purposes. Our empirical comparison with Naïve Bayes, decision trees, linear support vector machines (SVMs), random forest, long short-term memory (LSTM) neural networks, and other techniques, is quite conclusive. The Tsetlin Machine either performs on par with or outperforms all of the evaluated methods on both the 20 Newsgroups and IMDb datasets, as well as on a non-public clinical dataset. On average, the Tsetlin Machine delivers the best recall and precision scores across the datasets. Finally, our GPU implementation of the Tsetlin Machine executes 5 to 15 times faster than the CPU implementation, depending on the dataset. We thus believe that our novel approach can have a significant impact on a wide range of text analysis applications, forming a promising starting point for deeper natural language understanding with the Tsetlin Machine.

## Introduction

Understanding natural language text involves interpreting linguistic constructs at multiple levels of abstraction: words forming phrases, which interact to form sentences, which in turn are interweaved into paragraphs that carry implicit and explicit meaning (Norvig 1987; Zhang, Zhao, and LeCun 2015). Because of the complexity inherent in the formation of natural language, text understanding has traditionally been a difficult area for machine learning algorithms (Linell 1982). Medical text understanding is no exception, both due to the intricate nature of medical language, and due to the need for transparency, through human-interpretable procedures and outcomes (Berge, Granmo, and Tveit 2017; Y. Wang et al. 2017).

Although deep learning in the form of convolutional neural networks (CNN), recurrent neural networks (RNNs), and Long Short Term Memory (LSTM) recently has provided a leap ahead in text categorization accuracy (Zhang, Zhao, and LeCun 2015; Conneau et al. 2017; Schmidhuber 2015; Subramanian et al. 2018), this leap has come at the expense of interpretability and computational complexity (Miotto et al. 2017). Simpler techniques such as Naïve Bayes, logistic regression, decision trees, random forest, k-nearest neighbors (kNN), and Support Vector Machine (SVM) are still widely used, arguably because they are simple and efficient, yet, provide reasonable accuracy, in particular when data is limited (Y. Wang et al. 2017; Valdes et al. 2016).

**Challenges Addressed in This Paper:** Despite recent successes, natural language continues to challenge machine learning (Schütze, Manning, and Raghavan 2008; M. Wu et al. 2017).

First of all, realistic vocabularies can be surprisingly rich, leading to high-dimensional input spaces. Depending on the feature selection strategy, text categorization has to deal with thousands to tens of thousands (or even more) of features. Even though natural language text typically contains high volumes of duplicated words (e.g., stop words), few are irrelevant (Zipf 1932). Low frequency terms further carry considerable information that may be relevant for categorization (Joachims 1998). As a result, a vocabulary can be so diverse that documents covering the same category may not even share a single medium-frequency term (Joachims 2002). As an example, the Electronic Health Records (EHRs) dataset that we target in the present paper abounds with jargon, misspellings, acronyms, abbreviations, and a mixture of Latin, English, and mother tongue (Berge, Granmo, and Tveit 2017).

---
[*] The last three authors have been listed alphabetically.

Another challenge is the inherently complex composition of natural language (Miotto et al. 2017; Linell 1982). For example, determining whether a patient is allergic to drugs or not based on EHRs has proven to be very hard without introducing hand-crafted rules (Berge, Granmo, and Tveit 2017). Rather than relying on the additivity of individual features to discriminate between categories, full sentences, paragraphs and even complete records must be considered in context. Indeed, the temporal dynamics of EHRs is a strong confounding variable. Simultaneously, EHRs may also contain ambiguous considerations and speculations which must be treated as false positives despite their similarity to true positives on the individual feature level (Berge, Granmo, and Tveit 2017; Uzuner et al. 2011).

A third challenge is interpretability. While rule-based methods such as decision trees are particularly easy to understand, other techniques tend to be more complex (Y. Wang et al. 2017; Mittelstadt et al. 2016). Neural networks are arguably among the most complicated, and have been criticized for being "black boxes" (Miotto et al. 2017). That is, it is very difficult for humans to understand how they produce their results. Even trained data scientists may have problems explaining their outcomes, or such explanations may demand comprehensive and time-consuming analysis. It has been argued that as doctors or nurses are expected to justify their decisions, so should machine learning algorithms be able to explain *their* decisions (Y. Wang et al. 2017; Mittelstadt et al. 2016). Recent progress on post processing attempts to address this issue (M. Wu et al. 2017). However, such approaches add to the already inherent complexity of neural networks by introducing a new abstraction layer. They are further yet not capable of providing both an exhaustive and compact explanation of the reasoning going on inside a neural network.

**Paper Contributions:** In this paper, we introduce the first approach to text categorization that leverages the recently introduced *Tsetlin Machine* (Granmo 2018). The Tsetlin Machine has caught significant interest because it facilitates human understandable pattern recognition by composing patterns in propositional logic. Yet, it has outperformed state-of-the-art pattern recognition techniques in benchmarks involving pattern discrimination, image recognition, and optimal move prediction for board games (Granmo 2018).

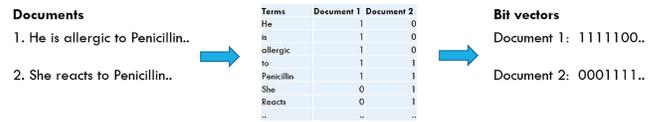

*Figure 1. Example of bit vector representation of document sentences, taken from the medical dataset used in one of the experiments.*

The Tsetlin Machine builds on the Tsetlin Automaton, a pioneering solution to the multi-armed bandit problem developed by M. L. Tsetlin in the Soviet Union in the late 1950s/early 1960s (Tsetlin 1961; Granmo 2018).

At the heart of the Tsetlin Machine we find a novel game theoretic scheme that orchestrates *decentralized teams* of Tsetlin Automata. The orchestration guides the Tsetlin Automata to jointly learn arbitrarily complex propositional formulae in conjunctive normal form, capturing the various facets of the patterns faced (Granmo 2018). Such formulae have turned out to be particularly suited for human interpretation, while still allowing complex non-linear patterns to be formed (T. Wang et al. 2017).

Apart from being simple to interpret, the Tsetlin Machine uses bits to represent both inputs, patterns and outputs. Recognition of patterns, in turn, merely involves decentralized manipulation of bits. This provides the Tsetlin Machine with an advantage computationally, compared to methods that rely on more complex mathematical modelling, in particular when implemented on a GPU.

**Paper Organization:** The paper is organized as follows. In Section 2, we provide the details of our Tsetlin Machine based approach to text categorization. We further demonstrate how the conjunctive clauses are formed to capture complex patterns in natural language. Then, in Section 3, we evaluate our framework empirically on both the 20 Newsgroups and the IMDB movie review (IMDb) datasets, as well as on a non-public clinical dataset, contrasted against selected state-of-the-art techniques. Finally, in Section 4, we conclude and provide pointers for further research, including contextual language models for the Tsetlin Machine.

## Text Categorization with the Tsetlin Machine

In this section, we first present our Tsetlin Machine based framework for text categorization. To highlight the key characteristics of the framework, we provide a running example from the medical domain – detection of Allergy in Electronic Health Records (EHRs). Thereafter, we present the learning procedure itself, demonstrating step-by-step how the Tsetlin Machine extracts sophisticated linguistic patterns from labelled documents and simultaneously combats overfitting.

## Representing Text as Propositional Variables

The Tsetlin Machine takes a vector of k propositional variables, $\mathbf{X} = [x_1, x_2, ..., x_k]$, as input, each taking the value 0 or 1 (or equivalently, False or True). As illustrated in Figure 1, we represent text (sentences, paragraphs, or in our case, documents) as follows. First, we form a vocabulary, $\mathbf{V}$, consisting of the unique terms found in the target text corpus, $\mathbf{D}$. A propositional variable, $x_o$, is then introduced to represent each term, $t_o \in \mathbf{V}$. This allows us to model a document, $D \in \mathbf{D}$, as a vector, $\mathbf{X}$, of $|\mathbf{V}|$ propositional variables (stored as a vector of bits). The vector tells us which terms are present and which are absent in $D$.

**Example:** For the topic Allergy, terms like "penicillin" and "reacts" would for instance be of importance, providing the propositional variables $x_{\text{"penicillin"}}$ and $x_{\text{"reacts"}}$.

The coarse text representation we use here removes the information on the ordering of terms. For more fine-grained modelling of language, a document can be broken down into paragraphs and sentences, as explored in the paper conclusion. More refined vocabularies can also be built by applying basic NLP-preprocessing techniques, such as case lowering, removing non-informative punctuation, tokenizing into n-grams, and taking advantage of sentence and paragraph boundaries (Kovačević et al. 2013).

## Linguistic Patterns in Conjunctive Normal Form

To build patterns that accurately capture categories, we compose propositional formulae in *conjunctive normal form*. These relate the propositional variables into compounds that trigger on evidence for or against a category.

Each compound, i.e., a conjunctive clause, is built by employing the conjunction operator on the propositional variables and their negations (referred to as *literals*):

$$C_j(\mathbf{X}) = x_{q_1} \wedge x_{q_2} \wedge ... \wedge x_{q_r} \wedge \neg x_{q_{r+1}} \wedge ... \wedge \neg x_{q_s}.$$

Here, $q_u$ are indexes from $\{1, ..., k\}$ that identify which propositional variables take part in the conjunction, as well as their role (negated or unnegated).

**Example:** For instance, the clause "rash" ∧ "reaction" ∧ "penicillin", can potentially act as evidence (a vote) for the category Allergy.

The beauty of the conjunctive normal form is that it can capture arbitrarily refined patterns by looking at multiple terms in conjunction, forming a global view of the input vector. This as opposed to modelling features independently, such as in linear models like the Naïve Bayes Classifier.

Finally, note the similarity between the conjunctive clauses in our model and the Boolean model of information retrieval, where Boolean expressions of terms are used to form queries by means of the AND-, OR-, and NOT operators (Schütze, Manning, and Raghavan 2008).

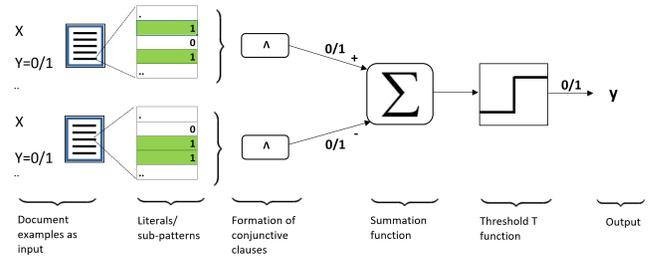

*Figure 2. The Tsetlin Machine architecture, introducing clause polarity, a summation operator collecting "votes", and a threshold function arbitrating the final output.*

## Categorization: Adding up Evidence from Clauses

As illustrated in Figure 2, in order to provide a rich and robust representation of each category, a Tsetlin Machine utilizes multiple clauses, $\underline{\mathbf{C}} = \{C_1, ..., C_m\}$. Each clause is further assigned a polarity (+/-). We use odd indexed clauses to capture patterns that provide evidence on the *presence* of a category (positive polarity), while even indexed clauses capture evidence on the *absence* of the category (negative polarity).

As further depicted in Figure 2, evidence obtained from clauses are aggregated by summation:

$$f_\Sigma(X) = \left(\sum_{j=1,3,...}^{m-1} C_j(X)\right) - \left(\sum_{j=2,4,...}^{m} C_j(X)\right)$$

The concluding categorization is finally decided from the sign of $f_\Sigma(X)$:

$$y = 1 \text{ if } f_\Sigma(X) > 0 \text{ otherwise } 0.$$

That is, only when the number of clauses providing a positive output outweigh those with negative output is the document assigned the target category ($y = 1$, i.e., Allergy in our case).

Altogether, this means that the Tsetlin Machine both learns what a category looks like, and what it *does not* look like. By summing evidence for and against the category, the threshold mechanism arbitrates the final decision.

**Example:** Recall our clause: "rash" ∧ "reaction" ∧ "penicillin". If the terms "rash", "reaction", and "penicillin" are all present in the document being categorized, the clause would evaluate to 1. This outcome would count as evidence for Allergy if the clause has positive polarity, and if none of the clauses with negative polarity evaluates to 1, the document would be assigned the category Allergy ($y = 1$).

## Learning the Composition of Clauses

We have now presented the structure of our framework for text categorization. However, a critical part remains. Learning propositional formula from data has been notoriously difficult (Valiant 1984), with the space of candidate formulae growing exponentially with the number of propositional variables involved (Granmo 2018).

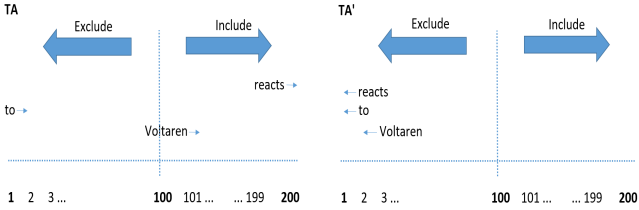

*Figure 3. Six Tsetlin Automata with 100 states per action. Each automaton learns to either Exclude or Include a candidate literal (a term or its negation) in a clause.*

This exponential growth becomes particularly severe in text categorization involving extensive vocabularies. A text categorization problem may consist of thousands, or even tens of thousands, of propositional variables. The full range of possible propositional formulae is thus immense. As a result, extreme care must be taken during training to maximize the precision and recall of the categorization, while simultaneously combating overfitting.

The pioneering nature of the Tsetlin Machine for text categorization can be summarized as follows. The Tsetlin Machine addresses the above challenges by decomposing the problem into a large number of simple decisions. These decisions are coordinated by a novel decentralized *game* between independent Tsetlin Automata – one automaton per clause, per literal (Granmo 2018). Each automaton decides whether to Include or Exclude the assigned candidate literal in the given clause. That is, whether a literal should take part in a clause is collectively regulated by the competing presence of multiple candidate literals.

This learning is performed on-line, one training example, $(\widehat{X}, \widehat{y})$ at a time. The input vector $\mathbf{X} = [x_1, x_2, …, x_o]$ dictates which terms are present and which are absent in the current document. The target, $\hat{y}$, to be predicted is simply the category of the document (Allergy or No Allergy in our case).

**The Tsetlin Automaton:** The Tsetlin Automaton that we use can be defined as a quintuple (Thathachar and Narendra 1989): $\{\mathbf{\Phi}, \mathbf{\alpha}, \mathbf{\beta}, F(\cdot,\cdot), G(\cdot)\}$. $\mathbf{\Phi} = \{\phi_1, \phi_2, …, \phi_{2N}\}$ is the set of internal states. $\mathbf{\alpha} = \{\alpha_{Exclude}, \alpha_{Include}\}$ is the set of automaton actions. $\mathbf{\beta} = \{\beta_{Inaction}, \beta_{Penalty}, \beta_{Reward}\}$ is the set of inputs that can be given to the automaton. An output function $G[\phi_t]$ determines the next action performed by the automaton given the current automaton state: (1) $G[\phi_t] = \alpha_{Exclude}$ for $1 \leq t \leq N$; and (2) $G[\phi_t] = \alpha_{Include}$ for $N+1 \leq t \leq 2N$. Finally, a transition function $F[\phi_t, \beta_u]$ determines the new automaton state from: (1) the current automaton state and (2) the response of the environment to the action performed by the automaton. In all brevity, we have: (1) $\phi_{t+1} = F[\phi_t, \beta_{Penalty}]$ for $1 \leq t \leq N$; (2) $\phi_{t-1} = F[\phi_t, \beta_{Penalty}]$ for $N+1 \leq t \leq 2N$; (3) $\phi_{t-1} = F[\phi_t, \beta_{Reward}]$ for $1 < t \leq N$; $\phi_{t+1} = F[\phi_t, \beta_{Reward}]$ for $N+1 \leq t < 2N$; and (5) $\phi_t = F[\phi_t, \cdot]$ **otherwise** (Granmo 2018). The crucial issue is to design automata that can learn the optimal action when interacting with the environment.

**Example:** Figure 3 depicts six Tsetlin Automata with $N=100$ states per action, collaborating to build a single clause. The three Tsetlin Automata to the left (TA) control the terms "to", "Voltaren" and "reacts", while the three to the right (TA') control the negation of the same terms. Those moving away from the central states in the figure ("Voltaren", "reacts", ¬"reacts", ¬"to", ¬"Voltaren") have received a $\beta_{Reward}$, while those moving towards the center have received a $\beta_{Penalty}$.

| Document→ Target Clause evaluates to | | 1 | | 0 | |
|---|---|---|---|---|---|
| Target Literal evaluates to | | 1 | 0 | 1 | 0 |
| Include literal | P (Reward) | $\frac{s-1}{s}$ | NA | 0 | 0 |
| | P (Inaction) | $\frac{1}{s}$ | NA | $\frac{s-1}{s}$ | $\frac{s-1}{s}$ |
| | P (Penalty) | 0 | NA | $\frac{1}{s}$ | $\frac{1}{s}$ |
| Exclude literal | P (Reward) | 0 | $\frac{1}{s}$ | $\frac{1}{s}$ | $\frac{1}{s}$ |
| | P (Inaction) | $\frac{1}{s}$ | $\frac{s-1}{s}$ | $\frac{s-1}{s}$ | $\frac{s-1}{s}$ |
| | P (Penalty) | $\frac{s-1}{s}$ | 0 | 0 | 0 |

*Table 1. The Type I Feedback Table has been designed to combat false negative output. It is triggered when a document correctly evaluates to true (i.e., both output value and target value is 1), or wrongly to false (i.e., output value is 0, while target value is 1).*

**Remark:** Note that a state near 1 conveys very strong support for $\alpha_{Exclude}$, while a state close to 2N means strong support for $\alpha_{Include}$. The center reflects uncertainty.

**The Tsetlin Machine Game:** The whole team of Tsetlin Automata, across all the categories and clauses, is orchestrated by means of a novel game (Granmo 2018). In this game, the Tsetlin Automata partake as independent players. Two simple tables specify the complete multi-dimensional game matrix, where the Tsetlin Automata are the decision makers: (1) *Type I Feedback* and (2) *Type II Feedback*. Each round of the game plays out as follows:

1. The arrival of a labelled document $(\widehat{X}, \widehat{y})$ signals the start of new game round.
2. Each Tsetlin Automaton decides whether to *include* or *exclude* its designated literal, leading to a new configuration of clauses $\underline{\mathbf{C}}$. Recall that each decision is based on the state of the Tsetlin Automaton, as exemplified in Figure 3.
3. Each clause, $C_j \in \underline{\mathbf{C}}$, is then evaluated with $\widehat{X}$ as input.
4. The final output, y, of the Tsetlin Machine is decided by thresholding $f_\Sigma(X)$, compared with the target output $\hat{y}$.
5. Each Tsetlin Automaton is independently and randomly given either Reward, Inaction, Penalty feedback, according to Table 1 or Table 2:
    a. If $\hat{y} = 1$ (true positive or false negative) the *Type I Feedback Table* is activated, while the *Type II Feedback Table* is activated if $\hat{y} = 0$ and y = 1 (false positive).
    b. As seen in the tables, the probability of each kind of feedback is decided by the current action of the automaton (either include or exclude), the value of the target clause, $C_j$, and the value of its assigned literal $x_k / \neg x_k$.
6. After all the Tsetlin Automata has received feedback and updated their states accordingly, exemplified in Figure 3, a new round starts.

| Document → | Target Clause evaluates to | 1 | | 0 | |
|---|---|---|---|---|---|
| | Target Literal evaluates to | 1 | 0 | 1 | 0 |
| Include literal | P (Reward) | 0 | NA | 0 | 0 |
| | P (Inaction) | 1.0 | NA | 1.0 | 1.0 |
| | P (Penalty) | 0 | NA | 0 | 0 |
| Exclude literal | P (Reward) | 0 | 0 | 0 | 0 |
| | P (Inaction) | 1.0 | 0 | 1.0 | 1.0 |
| | P (Penalty) | 0 | 1.0 | 0 | 0 |

*Table 2. The Type II Feedback Table combats false positive output. It is triggered when a document has wrongly evaluated to false, i.e. output value is 1, while the target value is 0.*

As shown in Figure 3, a Tsetlin automaton that receives a Reward moves away from the center (the Tsetlin automaton becomes more confident in the action it currently selects). Conversely, when receiving a Penalty, the state of a Tsetlin Automaton shifts towards the center states, N/N+1 (100 or 101), signaling increased uncertainty.

The *Type I Feedback Table* is activated when a document from is either correctly assigned the target category (true positive), or mistakenly ignored (false negative). The feedback given by this table has two effects. Firstly, clauses are refined by introducing more literals from the document. Left alone, this effect will make the clause *memorize* the document, by including *every* literal. However, this effect is countered by the second effect. The second effect is weaker by a factor of $s$, seeking to make all clauses evaluate to 1, whenever a document of the target category appears (and the clause has positive polarity). This effect counters overfitting. Indeed, the "granularity" of the clauses produced can be finely controlled by means of $s$.

The second table, the *Type II Feedback Table*, is activated when a document is wrongly assigned the target category (a false positive categorization). The effect of this table is to introduce literals that render the clauses false whenever they face a document of the incorrect category.

Both of the tables are thus interacting, making the whole game of Tsetlin Automata converge towards robust pattern recognition configurations.

Note that in order to effectively utilize sparse pattern representation capacity (a constrained number of clauses), we use a threshold value $T$ as target for the summation $f_\Sigma(X)$. That is, the probability of activating *Type I Feedback* is: $(T - \max(-T, \min(T, f_\Sigma(X)))/2T$, while *Type II Feedback* is activated with probability: $(T + \max(-T, \min(T, f_\Sigma(X)))/2T$. If the votes accumulate to a total of +/- T or more, neither rewards or penalties are handed out to the involved Tsetlin Automata. The threshold mechanism thus helps to alleviate the vanishing signal-to-noise ratio problem, as it effectively removes clauses (and their literals) that already are able to capture a pattern (supporting the correct classification) from further feedback (Granmo 2018).

**Remark:** The computational simplicity and small memory footprint of Tsetlin Automata, combined with the decentralized nature of the above game, make the Tsetlin Machine particularly attractive for execution on GPUs. This is explored further in the empirical results section.

### Detailed Walkthrough of the Learning Steps

Figure 4 provides a step-by-step walkthrough that exemplifies how the Tsetlin Machine is able to gradually learn the target concept Allergy, building and refining clauses, document-by-document. For the sake of clarity, we focus on a substring of each document, despite the fact that the steps we now describe happen for each and every term of the vocabulary, for each document processed. We explain the figure from left to right, starting with Example 1 (from the top).

**Processing of Document 1:** The first document, $D_1$, contains the substring "reacts to Voltaren". Each of these terms are associated with propositional variables, say, $x_{29123}$, $x_{32232}$, and $x_{37372}$. Again, for clarity, we focus on two of the clauses associated with the target concept, $C_1(\mathbf{X})$ and $C_2(\mathbf{X})$, both with positive polarity. Each clause may potentially include any of the propositional variables, as well as their negation. Thus, we have two Tsetlin Automata for each term, for every clause. The first Tsetlin Automaton in the pair (referred to as TA) decides whether to Exclude or Include the propositional variable as is. The second one (referred to as TA') decides whether to Exclude or Include the *negation* of the variable.

Starting from the left, the columns TA and TA' list the state of each of the associated automata controlling $C_1(\mathbf{X})$, before they have been updated. As seen, the term "reacts" is already Included in $C_1(\mathbf{X})$ because the state of the corresponding automaton is greater than or equal to N+1 (i.e., 101). It is further included with confidence because the state is almost at the extreme end, i.e., 2N. The other two terms, "to" and "Voltaren", are excluded. "to" is excluded with confidence (state *1*) and "Voltaren" is excluded with uncertainty (state *100*). In other words, the clause takes the form $C_1(\mathbf{X}) = x_{29123}$, initially.

With $y=1$, document $D_1$ has been flagged as an example of the target category. Accordingly, the *Feedback Type I Table*, described previously, is activated as depicted in the figure, and the new updated states of the Tsetlin Automata are listed next. As seen, after the update, "Voltaren" is now also included in the clause, increasing categorization precision: $C_1(\mathbf{X}) = x_{29123} \wedge x_{37372}$.

**Processing of Document 3:** Further in the processing, another document, $D_3$, with target $y = 1$ comes along. This time $C_1(\mathbf{X})$ evaluates to $0$, because "Voltaren" is absent from $D_3$. However, another clause $C_2(\mathbf{X}) = x_{29123}$ happens to only include "reacts", and thus evaluates to *1*. This in turn, triggers *Feedback Type I* for clause $C_2(\mathbf{X})$, again leading to an update of the Tsetlin Automata states. As seen, the "Apocillin" is now included in the clause because the associated

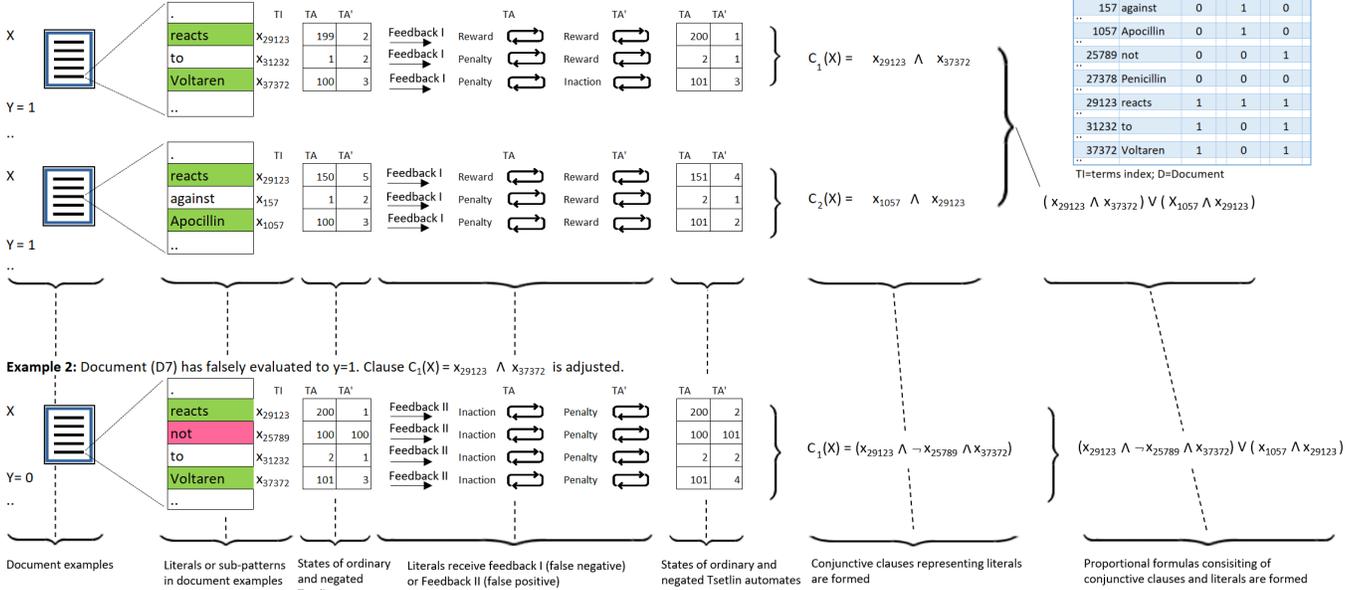

*Figure 4. Step-by-step example of Tsetlin Machine-based learning, demonstrating how clauses are composed and refined.*

Tsetlin Automaton changed action by going from state 100 to *101*: $C_2(\mathbf{X}) = x_{29123} \wedge x_{1057}$.

**Processing of Document 7:** Document $D_7$ is an example of a document not belonging to the category, with $y = 0$. This example highlights the difficulty of text categorization task at hand. Indeed, $C_1(\mathbf{X})$ evaluates to *1* because the terms "reacts" and "Voltaren" are present, leading to a false positive categorization. However, this activates Feedback Type II. This time, clause $C_1(\mathbf{X})$ is updated by attempting to modify the clause so that it evaluates to 0 instead. This is done by penalizing all exclude actions that would have led the clause to be evaluated 0, if one of them had been included instead. The most uncertain of those being penalized is the literal that negates "not". After the feedback, this literal is now being included in the clause: $C_1(\mathbf{X}) = x_{29123} \wedge x_{37372} \wedge \neg x_{25789}$. In this case, the false positive categorization was eliminated by this simple change. In general, however, the game needs to played for several epochs before all of the clauses find their final form.

## Experiments

In this section, we evaluate the Tsetlin Machine empirically, in comparison with selected text categorization techniques. To this end, we use two public datasets and one proprietary: 1) The 20 Newsgroups dataset (20 Newsgroups Data Set n.d.; Lang 1995); 2) The IMDb dataset (Maas et al. 2011); and 3) A clinical dataset with authentic EHRs from a hospital (Berge, Granmo, and Tveit 2017). The two first datasets facilitate comparison with previous research, while the third focuses on real-life performance in a challenging application area. The evaluation metrics we employ are macro-averaged accuracy, recall, precision, and F-measure. Results are presented as mean percentage, with 95% confidence intervals. Our CUDA and C source code for text classification with the Tsetlin Machine can be found at https://github.com/cair/TextUnderstandingTsetlinMachine.

### Experimental Setup

The experiments were conducted on a server (Intel i7-8700K 12-core CPU, 2GB SSD, and 64GB of memory) with Ubuntu 18.04 LTS x64 installed. We adopted the Python (with Cython C/C++ extensions) or CUDA/GPU capable C++ version of the Multi-Class Tsetlin Machine to run the experiments. We used implementations from the Scikit-learn, Keras, and Tensorflow libraries to produce results for the IMDb dataset. This included Naïve Bayes, logistic regression, decision tree, Random forest, linear SVM, kNN, multilayer perceptron (MLP), LSTM, LSTM CNN, Bidirectional LSTM (Bi-LSTM) and Bi-LSTM CNN. For the 20 Newsgroups and the clinical dataset experiments we used Weka (v3.9), except for StarSpace (L. Wu et al. 2017). The classifiers we employed in Weka were Naïve Bayes, decision tree (J48/C4.5-based decision tree algorithm, Logistic regression (Multinomial logistic regression), Random forest, kNN (Instance-Based k - a nearest neighbors variant), SMO (Sequential Minimal Optimization - an SVM variant), and MLP (DL4jMlp). NLP-preprocessing in Weka and Python (NLTK and Scikit-learn libraries) was performed to replicate the described Tsetlin Machine configuration for each of the experiments. This involved running the StringToWordVector filter in Weka to perform cleaning, tokenization of the input data into unigrams and quadrigrams, and the InfoGainAttributeEval method to determine the most significant attributes. We used the authors' published training/test data split (50%-50%) for the experiments on the IMDb dataset (Maas et al. 2011). For the 20 Newsgroups and the clinical we performed 10-fold cross-validation, repeated 10 times to produce representative averages with small variance. An 80%-20% split was used for the 20 Newsgroups dataset, and a 90%-10% split was implemented for the clinical dataset.

### The 20 Newsgroups Dataset

The results for the modified[2] 20 Newsgroups dataset with 4 classes are reported in Table 1. The Tsetlin Machine we employ here was configured using 8006 selected features, 1600 clauses, and 100 states. Furthermore, we used an s-value of 8.0 and a Threshold T of 25. The Tsetlin automata was run for 100 epochs per fold.

As seen in Table 3, The Tsetlin Machine outperforms all of the approaches for text categorization we evaluated, apart from the recall of the Support Vector Machine (SVM), which was identical to the recall of the Tsetlin Machine. Note that a more or less identical dataset has previously been used to evaluate different Convolutional Neural Network architectures (CNN; CNN-topic, CNN-channel, CNN-model, CNN-concat) (Xu et al. 2016) producing accuracy scores in the range 94.8-95.5%.

| Method | Precision | Recall | F-measure |
|---|---|---|---|
| Naïve Bayes | 86.1±1.5 | 98±0.6 | 91.6±0.9 |
| Logistic regression | 99.3±0.4 | 99.6±0.3 | 99.4±0.2 |
| Decision tree | 99.6±0.2 | 98.9±0.4 | 99.2±0.2 |
| Random forest | 91±1.7 | 99.3±0.4 | 94.9±0.9 |
| kNN | 69.8±2.3 | 96.4±1.0 | 80.9±1.5 |
| SVM | 99.7±0.2 | 99.8±0.2 | 99.7±0.2 |
| MLP | 97.7±1.4 | 96.6±1.7 | 97.1±1.1 |
| Tsetlin Machine | 99.8±0.2 | 99.8±0.2 | 99.8±0.2 |

*Table 3. Average classification results for the 20 Newsgroup dataset.*

Above, accuracy was maximized by means of careful feature selection. To investigate scalability and robustness, however, we also performed experiments without feature selection for the Tsetlin Machine, giving a total of 25 840 features. Using respectively 5168 and 10 336 clauses (the other parameters were kept the same), we were able to maintain an average accuracy of 99.8% for both cases. Accordingly, the Tsetlin Machine appears to generalize well in high dimensional feature spaces, potentially eliminating the need for feature selection. This would make the application of text categorization considerably easier.

### The IMDb Dataset

Table 4 shows the results for the IMDb dataset. The Tsetlin Machine was here configured to use 5 000 selected features, 10 000 clauses, and 500 states. Furthermore, we used an s-value of 27.0, a Threshold T of 20, and ran the Tsetlin Machine for 200 epochs. Similar to the 20 Newsgroups dataset, the reported results are from the final epoch.

As observed in Table 4, the Tsetlin Machine again outperformed the other methods. The macro averaged accuracy of the Tsetlin Machine in this experiment was 89.2± 0.2, with 89.7% being the highest accuracy over all the epochs and folds. To facilitate a statistically robust comparison of the different techniques, the results in the table are mean scores from 100 independent runs, together with 95% confidence intervals. By reporting the mean, our results are by nature lower than what one achieves by e.g. reporting the 5[th] best accuracy over for instance 100 trials.

| Method | Precision | Recall | F-measure |
|---|---|---|---|
| Multinom. Naïve Bayes | 85.9±0.0 | 86.1±0.0 | 86.0±0.0 |
| Logistic regression | 86.5±0.0 | 87.6±0.0 | 87.1±0.0 |
| Decision tree | 71.1±0.0 | 68.4±0.0 | 69.7±0.0 |
| Random forest | 85.9±0.0 | 86.1±0.0 | 86.0±0.0 |
| kNN | 58.4±0.0 | 63.5±0.0 | 60.8±0.0 |
| Linear SVM | 88.0±0.0 | 89.1±0.0 | 88.5±0.0 |
| MLP | 82.0±0.0 | 84.0±0.0 | 83.0±0.0 |
| LSTM | 87.2±3.5 | 84.3±4.8 | 85.6±3.0 |
| LSTM CNN | 89.5±1.1 | 86.8±1.9 | 88.1±0.5 |
| Bi-LSTM | 87.6±4.1 | 83.9±5.4 | 85.5±2.8 |
| Bi-LSTM CNN | 88.3±1.2 | 87.5±2.4 | 87.9±0.9 |
| Tsetlin Machine | 90.1±1.0 | 88.2±1.5 | 89.1±0.4 |

*Table 4. Average classification results for the IMDb dataset.*

Utilizing a convolutional neural network (CNN) and a Long short-term memory (LSTM) network to classify the same data, Yenter and Verma (Yenter and Verma 2017) achieved maximum accuracy scores of between 89.22% and 89.5% on five different models, and thus concluded that their proposed model outperformed prior relevant research (Rahman, Mohammed, and Al Azad 2016; Tripathy, Agrawal, and Rath 2016; Maas et al. 2011). Recent years, several authors have also reported even better results by using ensemble methods or by introducing combinations of techniques such as entropy minimization loss, adversarial, and virtual adversarial training (S. Wang and Manning 2012; Mesnil et al. 2014; Zhao 2016; Johnson and Zhang 2015; Dieng et al. 2016; Miyato, Dai, and Goodfellow 2016; Sachan, Zaheer, and Salakhutdinov 2018; Radford, Jozefowicz, and Sutskever 2017; Gray, Radford, and Kingma 2017; Howard and Ruder 2018). For example, Wang and Manning reported an accuracy of 91.22% by implementing SVM with NB features (NBSVM-bi) (S. Wang and Manning 2012). Using deep learning, Miyato et al. (Miyato, Dai, and Goodfellow 2016) were able to achieve an accuracy score of 94.09% by using adversarial training methods. Finally, by implementing a mixed objective func-tion (L-mixed) and word embedding, Singh Sachan et al. accomplished an accuracy score of 95.68% (Sachan, Zaheer, and Salakhutdinov 2018).

---

[2] This dataset is derived from the original 20 Newsgroups dataset (the "19997" version) and contains 16 000 newsgroup documents. For comparison with Xu et al. (Xu et al. 2016), we strived to use a similar setup; i.e. the sub-categories has been grouped into the most common categories computers, recreation, science and politics.

In our further work, we intend to investigate how we can enhance our relatively plain modelling of text (presence and absence of terms) to further increase accuracy. One option is for example to introduce pretrained word embedding.

### The Clinical Dataset

Table 5 displays results for the clinical dataset. Besides reporting categorization accuracy results, we also report total run time in hours. The complexity of the dataset is reflected in the high standard deviations produced by all of the algorithms. To produce the final results presented here, the Tsetlin Machine was configured to use 38876 features, 500 clauses, 100 states, s-value of 3.0, Threshold T of 25, and to run 100 epochs per fold.

| Method | Precision | Recall | F-measure |
|---|---|---|---|
| Naïve Bayes | 65±7.2 | 76.2±9.1 | 69.9±6.6 |
| Logistic regression | 73.4±9.1 | 66.9±9.1 | 69.6±7.2 |
| Decision tree | 59.7±9.1 | 57.3±12.0 | 57.9±8.7 |
| Random forest | 63±9.3 | 56.1±10.3 | 58.9±8.4 |
| kNN | 68±14.6 | 36.8±10.8 | 47.1±11.4 |
| SVM | 69.3±8.4 | 59.7±10.0 | 63.7±8.0 |
| MLP | 70.3±17.6 | 67.1±12.3 | 67.2±10.9 |
| StarSpace | 61.7±13.2 | 59.8±9.7 | 59.3±9.0 |
| Tsetlin Machine | 70.3±10.8 | 70.2±11.6 | 69.3±8.1 |

*Table 5. Average classification results for the clinical dataset.*

When considering the F-measure results, the Tsetlin Machine on this classification task has to yield by a very small margin to Naïve Bayes in first place and Logistic regression in second place. Interestingly, these two linear algorithms outperform all of the other methods, including the usually more capable nonlinear algorithms (e.g., SVM, neural network, and the Tsetlin Machine). A peak accuracy of 79.1% was achieved in epoch 294 when running the Tsetlin Machine for an extended total of 2000 epochs, possibly reflecting temporary overfitting on the test set (in later epochs, the accuracy results stabilized around 70%). It is reasonable to believe that the Naive Bayes and Logistic regression classifiers in this experiment gain from their simple (linear) structure, preventing them from fitting training data too closely (Ng and Jordan 2002). Similar to Naïve Bayes, the Tsetlin Machine has also demonstrated accuracy advantages when faced with a lack of data, which to some extent may possibly explain why the Tsetlin Machine is able to outperform the other nonlinear capable algorithms employed in this experiment (Granmo 2018).

Note that we for all of the above experiments have also used a GPU implementation of the Tsetlin Machine. The GPU version executes 5 to 15 times faster than the CPU implementation, depending on the dataset.

Although we conducted extensive hyperparameter tuning for the Tsetlin Machine (manual grid search) and the other evaluated algorithms (e.g., autotuning by use of the CVParameterSelection metafilter in Weka), even better configurations may potentially exist. We have, however, strived to put an equal effort into optimizing each approach to facilitate a fair comparison.

## Conclusion and Future Work

This paper proposed a text categorization approach based on the recently introduced Tsetlin Machine. In all brevity, we represent the terms of a text as propositional variables. From these, we capture categories using simple propositional formulae that are easy to interpret for humans. The Tsetlin Machine learns these formulae from a labelled text, utilizing conjunctive clauses to represent the particular facets of each category. Our empirical results were quite conclusive. The Tsetlin Machine either performed on par with or outperformed all of the evaluated methods on both the Newsgroups and IMDb datasets, as well as on a non-public clinical dataset. On average, the Tsetlin Machine delivered the best recall and precision scores across the datasets.

In our further work, we plan to examine how to use the Tsetlin Machine for unsupervised learning of word embeddings. Furthermore, we will investigate how the sparse structure of documents can be taken advantage of to speed up learning. We further plan to leverage local context windows to learn structure in paragraphs and sentences for more precise text representation. Finally, based on our promising results we envision implementing the Tsetlin Machine in clinical decision support systems (Berge, Granmo, and Tveit 2017). In particular, we are interested in how structured medical data can be combined with the medical narrative to support precision medicine.

All in all, we believe that our novel Tsetlin Machine based approach can have a significant impact on a wide range of text analysis applications. Furthermore, we believe that the approach and results presented in this paper can form a promising starting point for deeper natural language understanding.